# UMono: Physical Model Informed Hybrid CNN-Transformer Framework for Underwater Monocular Depth Estimation

Jian Wang, Jing Wang, Shenghui Rong, Bo He, *Member, IEEE,*

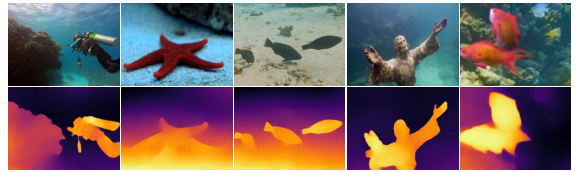

Fig. 1. Underwater images (first row) and corresponding depth maps (second row) estimated by the proposed UMono.

*Abstract*—Underwater monocular depth estimation serves as the foundation for tasks such as 3D reconstruction of underwater scenes. However, due to the influence of light and medium, the underwater environment undergoes a distinctive imaging process, which presents challenges in accurately estimating depth from a single image. The existing methods fail to consider the unique characteristics of underwater environments, leading to inadequate estimation results and limited generalization performance. Furthermore, underwater depth estimation requires extracting and fusing both local and global features, which is not fully explored in existing methods. In this paper, an end-to-end learning framework for underwater monocular depth estimation called UMono is presented, which incorporates underwater image formation model characteristics into network architecture, and effectively utilize both local and global features of underwater image. Specifically, UMono consists of an encoder with a hybrid architecture of CNN and Transformer and a decoder guided by medium transmission map. Firstly, we developed an Underwater Deep Feature Extraction (UDFE) block, which leverages CNN and Transformer in parallel to achieve comprehensive extraction of both local and global features. These features are effectively integrated via the proposed Local Global Feature Fusion (LGFF) module. By stacking the UDFE block as the basic unit, we constructed a hybrid encoder that generates four-stage hierarchical features. Subsequently, the medium transmission map is incorporated into the network as underwater domain knowledge, together with encoded hierarchical features are fed into Underwater Depth Information Aggregation (UDIA) module, which aggregates depth information from physical model and neural network by a proposed cross attention mechanism. Then, the aggregated features serve as the guiding information for each decoding stage, facilitating the model in achieving comprehensive scene understanding and precise depth estimation. The final estimated depth map is obtained through consecutive upsampling processing. Experimental results demonstrate that the proposed method is effective for underwater monocular depth estimation and outperforms the existing methods in both quantitative and qualitative analyses.

## I. Introduction

UNDERWATER monocular depth estimation aims to infer the three-dimensional information of the scene through sensors or algorithms, which is crucial for understanding underwater scenes and essential for marine tasks such as marine resource exploration [1], underwater robot navigation [2], [3] and 3D reconstruction [4] of underwater scenes. However, due to the influence of the underwater environment (such as water pressure, communication difficulties and energy supply limitations), deploying depth sensors underwater requires exceedingly stringent conditions while entails a substantially high cost. Besides, underwater images often exhibit characteristics [5] such as blur distortion, aberration, and feature degradation due to the influence of light conditions, suspended particles, and water quality, making it extremely difficult to obtain accurate depth maps through algorithmic processing. Consequently, underwater monocular depth estimation is a difficult and challenging task.

Traditional underwater monocular depth estimation approaches can be categorized into active methods and passive methods. Active methods utilize cameras [6] or sensors [7] to acquire depth information of a scene based on the optical properties. For example, structured light sensors [8] capture the deformation of projected structured patterns to infer the depth information of object surfaces; time-of-flight cameras [9] calculate scene depth information based on the round-trip time of light. However, the wavelength-dependent light attenuation [10] in underwater environments causes significant changes in optical physics properties, which severely affects the effective operation of sensors and leads to inaccurate monocular depth estimation results. In addition, the deployment of hardware devices underwater presents formidable challenges [11], notably the necessity for impeccably waterproof sealing, making these methods exceptionally hard to implement. Passive methods obtain the scene depth by inference underwater image formation model inversely, with some prior knowledge as known conditions. For example, Dark Channel Prior (DCP) [12] and its variations [13], [14], [15] tailored for underwater environments are widely applied to acquire scene depth to further restore underwater images. However, the prior is typically designed for specific situation

Jian Wang, Shenghui Rong and Bo He are with the School of Information Science and Engineering, Ocean University of China, Qingdao 266100, China (e-mail:wangjian2528@stu.ouc.edu.cn; rsh@ouc.edu.cn; bhe@ouc.edu.cn).

Jing Wang is with the Department of Agriculture and Fisheries, Queensland Government, Brisbane, QLD 4102, Australia.

and perform unsatisfactorily in diverse types of underwater images, thus resulting in poor generalization capability of these prior-based methods.

In recent years, deep learning-based methods have achieved significant advancements in various visual tasks, including monocular depth estimation [16], [17]. Researchers treat monocular depth estimation as a regression problem and exploit deep neural networks to extract depth-relevant features from RGB images in order to estimate the depth map from a single image. These methods benefit from the feature extraction capability of encoders and require training on a large-scale dataset, such as KITTI [18] and NYUDepth [19], to achieve superior accuracy and generalization capability. In the field of underwater studies, due to the lack of large-scale underwater depth datasets, researchers resort to unsupervised learning methods or utilize synthetic datasets to train models. For instance, UW-Net [20] proposed an unsupervised framework for monocular depth estimation based on cycle-consistent learning, UWGAN [21] trained generative adversarial networks (GAN) [22] using synthetic underwater datasets and presented a framework applicable for underwater monocular depth estimation and image enhancement. However, these methods rely on stacked convolution blocks and do not fully exploit the global features of underwater images. Additionally, they fail to consider the characteristics of underwater environment within the model framework, which hinders a comprehensive understanding of underwater scenes, thus leading to poor accuracy and generalization capabilities.

Considering the aforementioned issues, this paper proposes a novel end-to-end framework for underwater monocular depth estimation, termed as UMono. The key idea of the proposed approach lies in enhancing the representation ability of encoded features and integrating underwater domain knowledge into the model framework.

The proposed UMono consists of a hybrid encoder of CNN and Transformer and a decoder guided by medium transmission map. The encoder is composed of a stack of Underwater Depth Feature Extraction(UDFE) blocks, which leverages CNN and Transformer in parallel to extract features simultaneously. The former is used for extracting local features and spatial structural information, while the latter is utilized for modeling long-distance dependencies. The combination can ensure the continuity of depth values and the accuracy of overall distribution, which is crucial for underwater monocular depth estimation regression tasks. Based on the consideration of the varying dependence of depth values on local and global features. we propose a Local-Global Feature Fusion(LGFF) module for fusing local and global features, which efficiently combines extracted features by calculating weight maps for different features. Furthermore, through the stacking of Underwater Depth Feature Extraction(UDFE) block, an encoder capable of generating four-stage hierarchical features is constructed. Furthermore, underwater domain knowledge is integrated into the decoder network to achieve a more comprehensive understanding of underwater scenarios. The medium transmission map is a key parameter in the underwater image formation model, which represents the percentage of scene radiance reaching the camera and can reflect the depth information of the scene. Therefore, we propose to guide the decoding process with the medium transmission map and hierarchical encoded features, both of which contain crucial depth-related information and are highly beneficial for reconstructing the depth map. Specifically, the Underwater Depth Information Aggregation (UDIA) module is designed based on a cross attention mechanism, which adaptively integrates medium transmission map and hierarchical encoded features to acquire enhanced depth-related features for progressively guiding the decoding process.

The contributions of the proposed method can be summarized as follows.

1) A hybrid Underwater Depth Feature Extraction (UDFE) block that utilizes CNN and Transformer is proposed to extract and fuse local and global features required for underwater monocular depth estimation, ensuring the local details and global layout in the final depth map.
2) The underwater domain knowledge is incorporated into the decoding network by utilizing medium transmission map, fusing with hierarchical encoded features through a cross attention based Underwater Depth Information Aggregation(UDIA) module, guiding the decoding process for better understanding of underwater scenes.
3) The proposed UMono demonstrates effective performance in underwater depth estimation, achieving comparatively favorable results on benchmark datasets in terms of qualitative analysis and quantitative metrics.

The remaining sections of this paper are organized as follows. Section II introduces a brief review of related works for underwater monocular depth estimation. Section III describes the architecture of the proposed framework in detail. Section IV includes the qualitative, quantitative and ablation experiments results and analysis. Section V presents the conclusion of this paper.

## II. RELATED WORK

Traditional approaches attempt to estimating the medium transmission map to acquire depth information of the scene, which is then applied to the underwater image formation model for image restoration [23], [24]. These approaches predominantly depend on prior information or hand-crafted features as depth cues for scene depth estimation. Peng et al. [25] estimated scene depth based on the object blurriness, which employed the max filter and closing by morphological reconstruction on pixel blurriness map to generate and refine the depth map. Drews et al. [26] proposed a method for underwater image restoration and depth estimation by applying Underwater DCP, which improved DCP to adapt to the underwater environment. Peng et al. [27] extended their previous work [25], which utilized image blurriness and light absorption and estimated depth map more accurately. Peng et al. [28] estimated depth based on the depth-dependent color change, which first estimated a rough depth scattering by considering the impact of scattering on smoothness and gradients, and the refined the depth map via regression analysis on image intensity and depth. Chang et al. [29] proposed Submerged DCP for their simplified optics-based underwater image formation model and straightforwardly estimated the scene depth map. Raihan et al. [30] utilized background neutralization to acquire blurriness and background light information from single view image, as the depth cues to estimate depth map.

However, the inadequacy of these methods is apparent, as the prior information is only applicable to specific images, resulting in poor generalization capability of traditional approaches. Furthermore, due to the unique optical properties underwater, estimating underwater imaging parameters poses significant challenges, and traditional methods are also unsatisfactory in terms of accuracy.

In recent years, the rapid development of deep learning has shown tremendous potential in visual tasks, demonstrating great prowess and promise. Various methods have been proposed to promote the development of underwater monocular depth estimation through deep neural network. Gupta et al. [20] proposed an unsupervised method, they indirectly estimated depth map in the style transfer manner through cycle-consistent learning, which mapped the unpaired terrestrial hazy images and underwater images by using haze as the depth cue. Ye et al. [31] performed underwater color correction and depth estimation based on a joint learning architecture, achieving favorable results by exploiting the correlation between visual tasks. Hambarde et al. [21] proposed a GAN-Based method for underwater single image depth estimation and enhancement, the method employed the cascaded-stream architecture which estimated the fine-level depth by optimizing coarse-level depth. Zhao et al. [32] proposed a joint framework to synthesize underwater images and estimate depth based on adversarial learning, and designed a depth loss to mitigate texture leakage problems. Yang et al. [33] proposed a self-supervised framework for underwater monocular depth estimation, which considered underwater light attenuation as potential depth clues to estimate depth and utilized optical flow to refine the estimated depth map. Wang et al. [34] proposed a self-supervised model for underwater monocular sequences depth estimation which leveraged the correlation between consecutive frames to solve the scale ambiguity problem, the predicted depth was further utilized for underwater image enhancement. Liu et al. [35] combined single-beam echosounder with monocular camera, representing the single range measurement as mask and integrating it as additional cues for depth estimation. Amitai et al. [36] utilized specific underwater data augmentation and incorporated photometric prior into the loss for self-supervised learning. Yu et al. [37] designed a new input space for depth estimation based on of underwater domain characteristics and proposed a lightweight pipeline which combined MobileNetV2 [38] and MiniViT [39] for fast monocular depth estimation. Ebner et al. [40] generated sparse depth prior by extracting feature points, and then fed it into the network via dense parameterization, as an extension of [37], this method improved the prediction accuracy on dense depth estimation.

These underwater monocular depth estimation models primarily focus on transferring terrestrial models, lacking thorough consideration for the characteristics of underwater scenes. Furthermore, owing to extracting features via stacked convolution layers, current methods have limited capability in fully exploiting local and global information which are both essential for depth estimation. Thus, we are of the opinion that, a comprehensive extraction of features and integration of underwater domain knowledge are crucial for underwater monocular depth estimation.

## III. PROPOSED METHOD

The proposed framework for underwater monocular depth estimation is described in this section. Fig. 2 presents the overview of the proposed UMono, comprising an encoder utilizing hybrid CNN and Transformer, along with a decoder guided by the medium transmission map.

### A. Hybrid Encoder

Local information provides specific details and structure regarding the surfaces of nearby objects, which helps the recovery of continuous depth values within localized regions like object interiors and edges. Global information contributes to structure the entire scene comprehensively, which assists in enhancing the understanding of the scene and inferring the overall depth distribution [41]. Thus, integrating both local and global information can lead to the achievement of more accurate and robust underwater monocular depth estimation. CNN frequently employs $3 \times 3$ convolution kernels, restricting information aggregation to local regions and making it challenging to model long-range correlations [42]. In contrast,, Vision Transformer [43] leverages the self-attention mechanism to effectively extract global information. However, due to the patch embedding operation with relatively larger projection kernels, it may result in the loss of local details [44]. Thus, by considering the complementary properties of CNN and Transformer, we propose a hybrid encoder with the capability of extracting features with powerful representation.

Given an underwater image $I_{RGB} \in \mathbb{R}^{H \times W \times 3}$, the proposed encoder can generate four hierarchical features with scales of [1/4, 1/8, 1/16, 1/32] and channels of [$C_1$, $C_2$, $C_3$, $C_4$], documented as $E1$, $E2$, $E3$, and $E4$ respectively. Specifically, in each encoding stage, the input (image or feature maps) is first divided into patches of size $4 \times 4$. Subsequently, these patches are fed into [$N_1$, $N_2$, $N_3$, $N_4$] Underwater Depth Feature Extract blocks according to encoding stages to comprehensively extracting Global Information and Local Information. Furthermore, the hierarchical features of the stage are obtained through patch merging processing.

The Underwater Depth Feature Extraction block serves as a key component of the proposed encoder, which consists of two parts: feature extraction utilizing CNN block and Transformer block, along with feature aggregation facilitated by Local-Global Feature Fusion Module.

*1) Feature Extraction:* The CNN block is implemented by Depth-wise Separable Convolution [45], which employs depthwise convolution and pointwise convolution to capture spatial and cross-channel correlations, respectively. This factorized architecture achieves richer representations capacity while exhibits high efficiency, which can be formulated as

$$F_L = X + PWConv(BN(DWConv(X))) \quad (1)$$

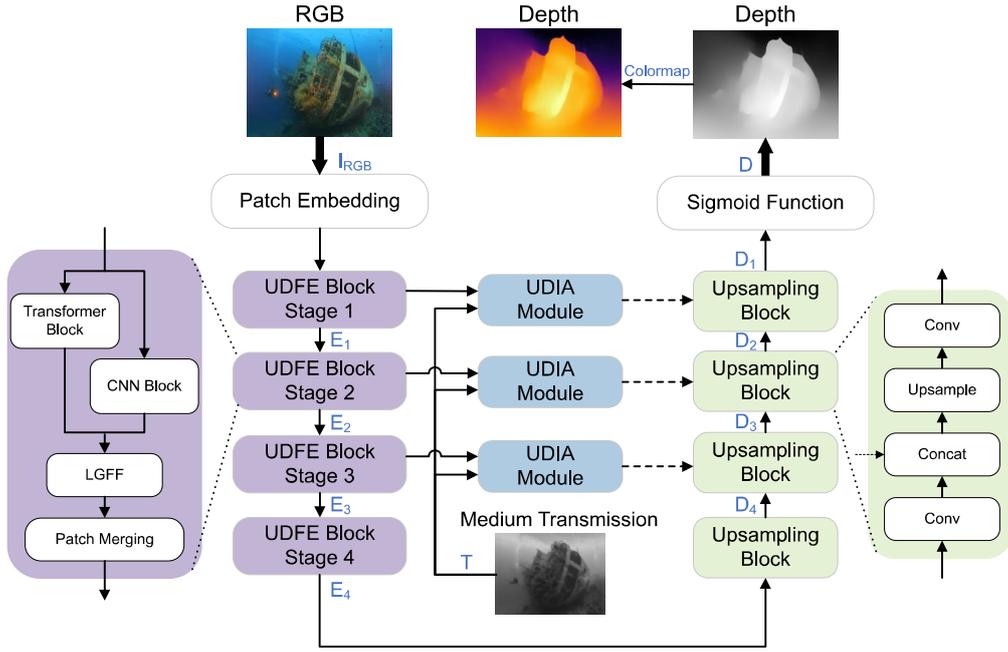

Fig. 2. Overview of the proposed UMono. The architecture of CNN block and Transformer block, LGFF module and UDIA module are detailed in Fig. 3, Fig. 4 and Fig. 6 respectively.

where $PWConv$ denotes the $1 \times 1$ pointwise convolution, $BN$ represents batch normalization operation and $DWConv$ stands for the $3 \times 3$ depth-wise convolution. The CNN block possesses the capability to effectively learn local features of an image, including edges, corners and clues related to object shapes, which are crucial for monocular depth estimation.

The Transformer block utilizes the spatial reduction scheme introduced in [46], [47] to efficiently compute the self-attention. Given an input feature $X \in \mathbb{R}^{H \times W \times C}$, the queries $Q^E = XW_q$, keys $K^E = XW_k$, and values $V^E = XW_v$ are obtained through linear projection, where $W_q$, $W_q$, and $W_q$ are linear projection matrices. $Q^E, K^E, V^E$ have the same dimensions $\mathbb{R}^{(HW) \times C}$, then the spatial reduction scheme is implemented to $K^E$ and $V^E$ for efficient calculation, which reshape $K^E$ and $V^E$ to dimensions $\frac{HW}{R}C$ via reducing the spatial dimension, where $R$ is the reduction ratio. The self-attention is calculated as

$$Attention(Q^E, K^E, V^E) = Softmax(\frac{(Q^E K^E)^T}{\sqrt{d_{head}}})V^E \quad (2)$$

where $d_{head}$ is the channel dimension of each head. Ultimately, the final global feature $F_G$ is achieved by concatenating the Attention calculated by each attention head:

$$F_G = Concat(A_1, \ldots, A_i, \ldots, A_N)W_g \quad (3)$$

where $Concat$ denotes the concatenation operation, $A_i$ represents the attention of $i$-th head, $N$ is the total head number, and $W_g$ is the linear projection matrix. The Transformer block utilizes spatial reduction attention, reducing the computational complexity by $R^2$ times. This approach enables the model to effectively understand the global structure and contextual information of the image, which is beneficial to achieving accurate global layout of depth in underwater scenes.

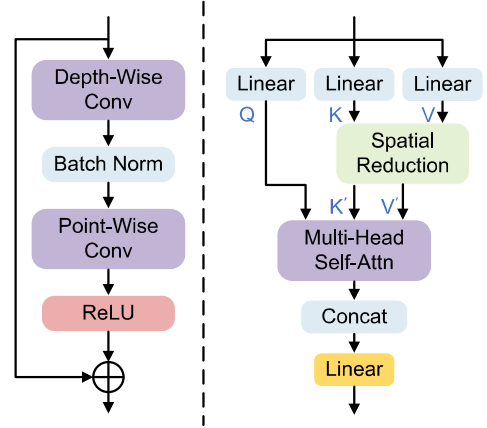

Fig. 3. Architecture of CNN and Transformer Block.

*2) Feature Aggregation:* Considering the varying dependence of depth values across pixels on both local and global information, for instance, the depth of edges and interiors of objects is primarily influenced by local regions to maintain consistency. Conversely, Certain pixels play a crucial role in maintaining a coherent depth layout, necessitating a higher reliance on global information. Therefore, we propose a Local Global Feature Fusion (LGFF) module designed to adaptively fuse local and global features based on their respective dependency levels.

The architecture of the proposed LGFF module is depicted in Fig. 4. Given the local feature $F_L \in \mathbb{R}^{H \times W \times C}$ and global feature $F_G \in \mathbb{R}^{H \times W \times C}$ extracted by CNN and Transformer, the LGFF first concatenates them along the channel dimension. The concatenated feature $F \in \mathbb{R}^{H \times W \times 2C}$ is then processed through multiple convolutional layers, including a sequence of

REPLACE THIS LINE WITH YOUR MANUSCRIPT ID NUMBER (DOUBLE-CLICK HERE TO EDIT) 5
truex3×3 Convolution, batch normalization, and ReLU function. Finally, a single-channel weight map $W \in \mathbb{R}^{H \times W \times 1}$ is obtained using the Sigmoid function. These weights are applied to $F_L$ and $F_G$ to construct the fused feature $F_E \in \mathbb{R}^{H \times W \times C}$ with a powerful representation. To sum up, The proposed LGFF can be expressed as

$$F = Concat(F_L, F_G) \quad (4)$$
$$W = Sigmoid(conv.b.r(F)) \quad (5)$$
$$F_E = F_L \otimes W + F_G \otimes (1 - W) \quad (6)$$

where $\otimes$ denotes Hadamard product. The LGFF effectively integrates local and global information, which is crucial for underwater monocular depth estimation, leading to more accurate results.

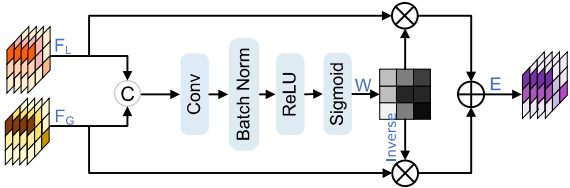

Fig. 4. Architecture of Local-Global Feature Fusion (LGFF) module.

## B. Medium Transmission Guided Decoder

The widely used underwater image formation model [48] can be mathematically expressed as

$$I_c(x) = J(x)T(x) + A(1 - T(x)) \quad (7)$$

where $I$ is the observed underwater image, $J$ is the restored underwater image, $c \in R, G, B$ and $x$ is the pixel index, $A$ is the ambient light, $T$ is the medium transmission map, which is related to depth of underwater scene and can be formulated as

$$T = e^{-\beta d} \quad (8)$$

where $\beta$ is the non-negative attenuation coefficient of water and $d$ is the scene depth. It is obviously that the medium transmission rate decreases exponentially with increasing depth values. Fig. 5 illustrates several RGB images, medium transmission maps, and depth maps, clearly demonstrating the inverse correlation between the transmission map and the depth map.

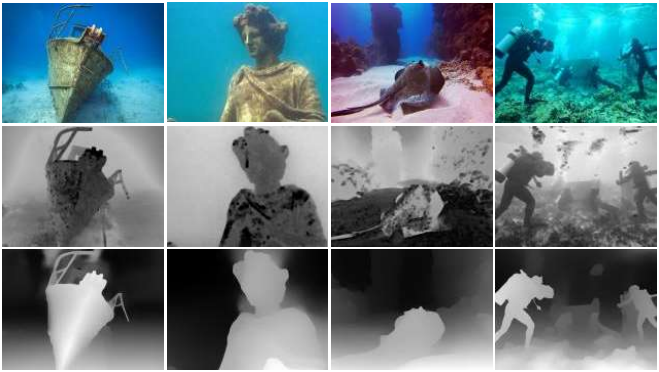

Fig. 5. Underwater images (first row), corresponding medium transmission maps (second row) and corresponding depth maps (third row).

Generally, the medium transmission map is pivotal for understanding underwater imaging process and is closely related to the depth, making beneficial to underwater monocular depth estimation task. Thus, we incorporate the medium transmission map as underwater domain knowledge into the decoding process, leading to a comprehensive understanding of underwater scenes and accurate depth estimation results. Specifically, we firstly estimated the medium transmission map via UDCP [13]. Then, we utilize the Underwater Depth Information Aggregation module to integrate the medium transmission map with hierarchical encoding features. Ultimately, each decoding stage takes corresponding integrated features as guidance to generate the final depth map.

The medium transmission map is estimated via prior-based method UDCP, which can be expressed as

$$T = 1 - \min_{y \in \Omega(x)} \left( \min_{c \in G, B} \frac{I_c(y)}{A} \right) \quad (9)$$

where $T$ is the estimated medium transmission map, $\Omega$ is a local patch centered at pixel $x$.

The medium transmission map contains depth-related information derived from the physical model of underwater imaging, while the hierarchical encoded features encompass depth-related information extracted by the deep neural network. The complementary information between these is crucial for enhancing the robustness of the model. Therefore, we propose an Underwater Depth Information Aggregation module designed to enhance depth information derived from the physical model and deep neural network.

The architecture of the proposed UDIA module is illustrated in Fig. 6. Given an encoded feature $E \in \mathbb{R}^{H \times W \times C}$ and medium transmission map $T \in \mathbb{R}^{H \times W \times 1}$, UDIA firstly apply the inverse operation to $\overline{T} = 1 - T$ to maintain a positive correlation with the depth value. To make full use of complementary information from $\overline{T}$ and $E$, a cross attention mechanism is designed to better guide the decoding process. Specifically, the learnable matrices $W_q$, $W_k$, and $W_v$ are implemented to project $\overline{T}$ and $E$ to query $Q_i^D$, key $K_i^D$, and value $V_i^D$:

$$\{Q_i^D, K_i^D, V_i^D\} = \{\overline{T}_i W_q, E_i W_k, E_i W_v\} \quad (10)$$

where $i$ denotes the $i$-th decoding stage. The complementary between information from physical models and neural networks is calculated as

$$I_i^C = Softmax(-\frac{(Q_i^D K_i^D)^T}{\sqrt{d_{head}}}) V_i^D \quad (11)$$

where $d$ is the dimension of the input feature, the application of $Softmax$ activation function to negative values is utilized to calculate the complementarity between query and key. Then $I_i^C$ is used to enhance the encoded feature by

$$F_i = E_i + Norm(I_i^C) \quad (12)$$

The integrated features encompass rich depth information which serves as guidance for the decoding process. In each decoding stage, the integrated feature $F_i$ is firstly concatenated with the decoding feature $D_i$ along the channel dimension. Subsequently, through conv-bn-relu and upsampling processes, the input decoding feature for the next decoding stage is derived. This process can be summarized as

$$D_{i-1} = UP(conv.b.r(Concat(F_i, D_i)))  \quad (13)$$

where $UP$ represents the bilinear upsample operation, $Concat$ denotes concatenation operation. After performing sigmoid and upsample processing on $D1$, we obtain the final predicted depth map $D \in \mathbb{R}^{H \times W \times 1}$.

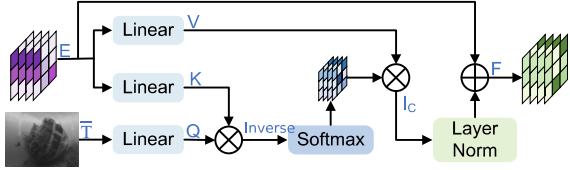

Fig. 6. Architecture of Underwater Depth Information Aggregation (UDIA) module.

### C. Loss Function

We use the combination of the $L_{l_2}$ loss and the Scale-Invariant Log loss $L_{SILog}$ [49] as the supervised loss function to train the proposed network, which can be defined as

$$L_{UMono} = \lambda L_2 + \mu L_{SILog}  \quad (14)$$

$$L_2 = \mathbb{E}\left[\left\|\hat{d}_i - d_i\right\|_2\right]  \quad (15)$$

$$L_{SILog} = \alpha \sqrt{\frac{1}{N}\sum_i^N g_i^2 - \frac{\beta}{N^2}\left(\sum_i^N g_i\right)^2}  \quad (16)$$

where $g_i$ = log $d_i$ - log $\hat{d}_i$ , $\hat{d}_i$ represents the ground truth depth, $d_i$ represents the predicted depth, $N$ is the total number of pixels in depth map. The balancing factor $\lambda$ and $\mu$ are set to 0.2 and 0.8 according to extensive experiments, $\alpha$ and $\beta$ are set to 10 and 0.85 same as [50].

## IV. EXPERIMENTS

In this section, we first introduce the implementation details, followed by a description of the experiment settings, including the Datasets, Evaluation Metrics, and Compared Methods. Then, we evaluate the performance of the proposed UMono through qualitative and quantitative analysis. Ultimately, ablation study is conducted to validate the effectiveness of the proposed architecture.

### A. Implementation Details

The proposed method is implemented in PyTorch [51] and trained on a single NVIDIA GeForce RTX 3090 with a batch size of 8. Adam [52] is utilized as the optimizer with weight decay of 1e-2. The initial learning rate is set to 1e-4 and the one-cycle learning rate strategy [53] is adopted with a poly adjustment schedule. The model is supervised by RGB-D images pairs covering diverse underwater scenes and trained for 80 epochs, which takes about 24 hours.

Specifically, the value of stacked UDFE blocks and embedding channels at each encoding stages are set to [3, 4, 6, 3] and [64, 128, 256, 512] respectively, the medium transmission maps are downsampled via pooling operations to adapt to different decoding stages.

### B. Experiment Settings

*1) Datasets:* The USOD10k dataset [54] contains 10255 underwater RGB images with corresponding depth maps from 12 different underwater scenes. To train and test the proposed UMono, we utilized the split 9229 training samples and 1026 testing samples in the experiments, each image has a resolution of $640 \times 480$.

The Atlantis dataset [55] is employed to evaluate the generalization capability of the UMono, which contains 3200 images with a resolution of $672 \times 512$ in various environment type.

*2) Evaluation Metrics:* The commonly used evaluation metrics [49] are adopted to evaluate the performance of the UMono, which are defined as follows

- **Abs Rel**: $\frac{1}{N}\sum_i^N \frac{|d_i - \hat{d}_i|}{\hat{d}_i}$
- **Sq Rel**: $\frac{1}{N}\sum_i^N \frac{\|d_i - \hat{d}_i\|^2}{\hat{d}_i}$
- **RMSE**: $\sqrt{\frac{1}{N}\sum_i^N \left\|d_i - \hat{d}_i\right\|^2}$
- **log10**: $\frac{1}{N}\sum_i^N \left|\log_{10} d_i - \log_{10} \hat{d}_i\right|$
- **Accuracy**:% of $d_i$ s.t. max $\left(\frac{d_i}{\hat{d}_i}, \frac{\hat{d}_i}{d_i}\right) = \delta < th$

where $d_i$ is the predicted depth, $\hat{d}_i$ is the ground truth, $N$ denotes the total number of pixels in depth map.

*3) Compared Methods:* We compared the proposed UMono with 7 methods, including GDCP [28], UW-Net [20], LapDepth [56], BTS [57], LiteMono [58], TransDepth [59], UDepth [37].

GDCP is a traditional prior-based method which measures scene depth to restore underwater images through image formation model. LapDepth and BTS are classic CNN-based methods for monocular depth method. LiteMono and TransDepth are typical and exemplary methods which implement a hybrid architecture of CNN and Transformer. UW-Net and UDepth are specifically designed for underwater monocular depth estimation, notably UW-Net is an unsupervised method based on CycleGAN [60] where UDepth employs a lightweight CNN and MiniViT for supervised learning.

### C. Qualitative Comparison

Qualitative comparisons of UMono with the representative methods on USOD10k dataset are shown in Fig. 7. GDCP [28] estimates erroneous results due to the limitations imposed by prior information, while UW-Net [20] also produces poor depth map with no haze in input images as depth cue. BTS [57] and LapDepth [56] fail to accurately capture depth distributions, resulting in deviations from the ground truth at both background and object. LiteMono [58] and TransDepth [59] estimate depth maps with better visual quality, while the results are not satisfactory on the detailed representation of objects, such as edge continuity and interior details. UDepth [37] performs well in depth layout but suffers from. In comparison, the proposed UMono effectively estimates precise depth maps exhibiting rich local details and accurate global layout, which suggests the satisfactory performance of the designed framework for underwater monocular depth estimation.

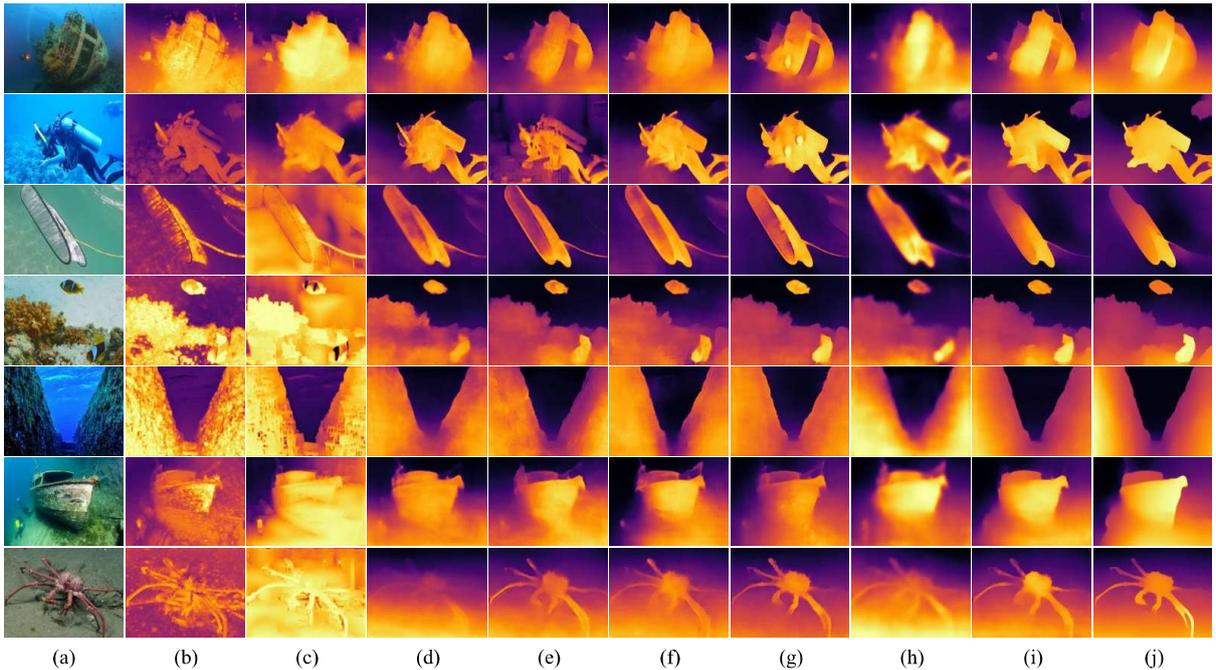

Fig. 7. Qualitative comparison results for depth estimation of the proposed UMono and compared methods on USOD10k dataset. (a) RGB. (b) GDCP [28]. (c) UW-Net [20]. (d) BTS [57]. (e) LapDepth [56]. (f) LiteMono [58]. (g) TransDepth [59]. (h) UDepth [37]. (i) UMono. (j) Ground Truth.

TABLE I
Quantitative comparison results for depth estimation of the proposed UMono and compared methods on USOD10k dataset. For $\delta_1$, $\delta_2$ and $\delta_3$, the higher value means the better performance. For other metrics, the lower value means the better performance. The value in bold and underline means the best and sub-optimal results respectively

| Method | $\delta_1$ | $\delta_2$ | $\delta_3$ | Abs Rel | Sq Rel | RMSE | log10 |
|---|---|---|---|---|---|---|---|
| GDCP [28] | 0.2100 | 0.3878 | 0.5376 | 1.6465 | 0.6501 | 0.2963 | 0.3444 |
| UW-Net [20] | 0.2276 | 0.3846 | 0.5101 | 2.5464 | 1.4091 | 0.3894 | 0.3753 |
| BTS [57] | 0.4242 | 0.6709 | 0.8010 | 0.6634 | 0.1154 | 0.1476 | 0.1876 |
| LapDepth [56] | <u>0.4664</u> | 0.6993 | 0.8238 | 0.5885 | 0.1005 | <u>0.1327</u> | <u>0.1709</u> |
| LiteMono [58] | 0.4629 | 0.6897 | 0.8111 | 0.5406 | 0.0940 | 0.1344 | 0.1811 |
| TransDepth [59] | 0.4636 | <u>0.7085</u> | <u>0.8267</u> | <u>0.5405</u> | <u>0.0910</u> | 0.1324 | 0.1780 |
| UDepth [37] | 0.4508 | 0.6723 | 0.7935 | 0.6810 | 0.1230 | 0.1430 | 0.1880 |
| UMono | **0.5237** | **0.7471** | **0.8519** | **0.4956** | **0.0778** | **0.1182** | **0.1526** |

Moreover, we also conduct qualitative comparisons on Atlantis dataset, as shown in Fig. 8, UMono still achieves scene understanding and yields accurate depth, which demonstrates the exemplary generalization capability owing to the effectiveness of the designed architecture.

### D. Quantitative Comparison

Quantitative comparisons on USOD10k dataset are conducted to quantify the performance of different compared methods, including GDCP [28], UW-Net [20], LapDepth [56], BTS [57], LiteMono [58], TransDepth [59], UDepth [37]. The results of commonly used evaluation metrics are demonstrated in Table I. Compared with other methods, UMono achieves the best performance for all metrics, with improvements of over 3.1% and 10.9 % in terms of $\delta_3$ and RMSE, respectively. Quantitative comparisons results on Atlantis dataset are presented in Table II. Among all the methods, UMono obtains the best performance on various scenes. The qualitative analysis results visually demonstrate the superiority and effectiveness of UMono for underwater monocular depth estimation.

### E. Ablation Study

Ablation experiments are designed to further analyze the effectiveness of the main components in UMono, including the Underwater Depth Feature Extraction (UDFE) block, Local-Global Feature Fusion (LGFF) module and Underwater Depth Information Aggregation (UDIA) module.

*1) The benefit of UDFE block:* The UDFE module is designed for leveraging CNN and Transformer in parallel to extract both local and global information for monocular depth estimation, in which the LGFF module effectively integrate the local and global features to enhance the representation

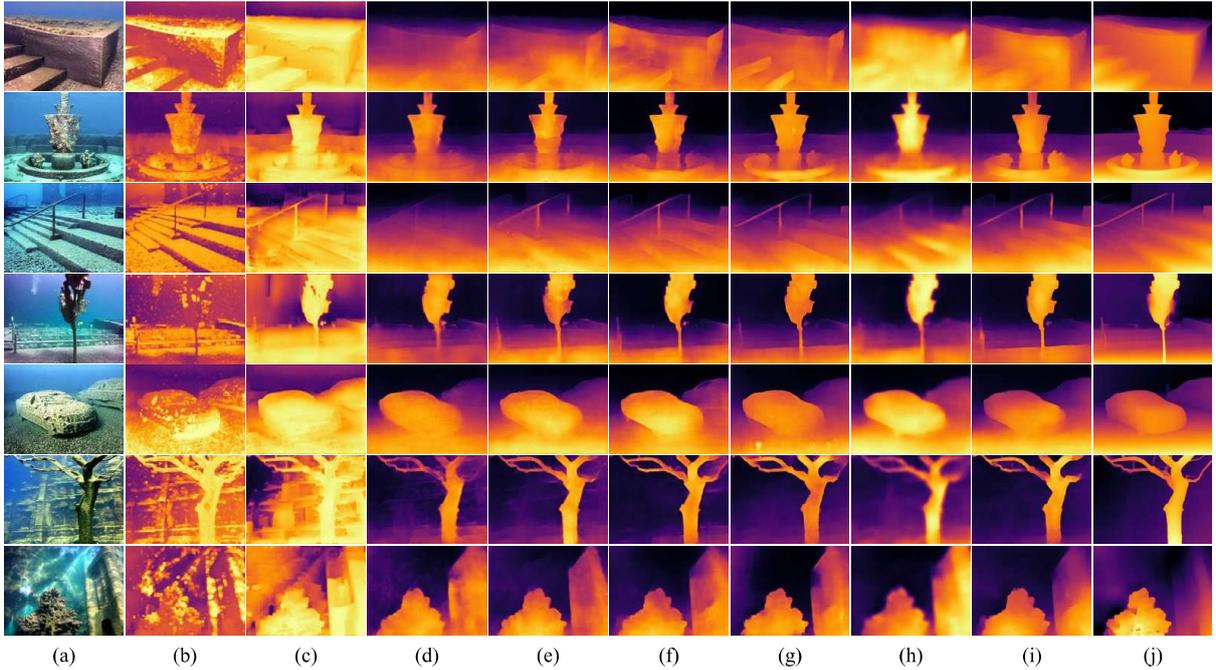

(a) (b) (c) (d) (e) (f) (g) (h) (i) (j)

Fig. 8. Qualitative comparison results for depth estimation of the proposed UMono and compared methods on Atlantis dataset. (a) RGB. (b) GDCP [28]. (c) UW-Net [20]. (d) BTS [57]. (e) LapDepth [56]. (f) LiteMono [58]. (g) TransDepth [59]. (h) UDepth [37]. (i) UMono. (j) Ground Truth.

TABLE II
Quantitative comparison results for depth estimation of the proposed UMono and compared methods on USOD10k dataset. For $\delta_1$, $\delta_2$ and $\delta_3$, the higher value means the better performance. For other metrics, the lower value means the better performance. The value in bold and underline means the best and suboptimal results respectively

| Method | $\delta_1$ | $\delta_2$ | $\delta_3$ | Abs Rel | Sq Rel | RMSE | log10 |
| --- | --- | --- | --- | --- | --- | --- | --- |
| GDCP [28] | 0.1636 | 0.3349 | 0.4762 | 3.3197 | 2.7447 | 0.4540 | 0.4108 |
| UW-Net [20] | 0.1489 | 0.2887 | 0.4302 | 1.6701 | 0.9291 | 0.4504 | 0.3610 |
| BTS [57] | 0.3753 | 0.6017 | 0.7366 | 1.1198 | 0.2188 | 0.1550 | 0.2260 |
| LapDepth [56] | 0.4062 | 0.6221 | 0.7501 | 1.0856 | 0.2041 | 0.1443 | 0.2144 |
| LiteMono [58] | 0.4047 | <u>0.6298</u> | 0.7545 | <u>0.8493</u> | <u>0.1573</u> | 0.1456 | 0.2117 |
| TransDepth [59] | <u>0.4078</u> | 0.6254 | <u>0.7602</u> | 0.8851 | 0.1580 | <u>0.1443</u> | <u>0.2060</u> |
| UDepth [37] | 0.3616 | 0.5632 | 0.6975 | 1.2664 | 0.3231 | 0.1799 | 0.2451 |
| UMono | **0.4096** | **0.6354** | **0.7694** | **0.8449** | **0.1450** | **0.1422** | **0.2017** |

capacity. We verify the effectiveness of UDFE by adjusting the architecture, including a total of 3 designs

- w/ CNN represents only utilizing CNN block for feature extraction.
- w/ Transformer represents only utilizing Transformer block for feature extraction.
- w/o LGFF represents replacing LGFF with the method of element-wise addition.

The ablation architecture are all train from scratch for 60 epochs with a batch size of 8. The qualitative and quantitative results are shown in Fig. 9 and Table III respectively.

As presented in Fig. 9, only utilizing CNN and Transformer block for feature extraction lead to relatively poor results, due to the lack or insufficiency of essential information necessary for monocular depth estimation. The results of 'w/o LGFF' also have an unsatisfactory performance on local details such as edges and fail to achieve the effective integration of features, which is further corroborated by the quantitative results in Table III. While the proposed UDFE with LGFF module based on the features dependency levels, effectively aggregates local and global information, leading to better local details and global layout, and an improvement of over 13.3% in term of Sq Rel. The experiments results demonstrate that the hybrid architecture is effective for feature extraction and the LGFF module enhances the features representation via feature aggregation, indicating the UDFE is beneficial for underwater monocular depth estimation.

TABLE III
QUANTITATIVE COMPARISON RESULTS OF ABLATION STUDY ON THE BENEFIT OF UDFE BLOCK. THE VALUE IN BOLD AND UNDERLINE MEANS THE BEST AND SUB-OPTIMAL RESULTS RESPECTIVELY

| Variant | $\delta_3$ | Abs Rel | Sq Rel | RMSE | log10 |
|---|---|---|---|---|---|
| w/ CNN | 0.7834 | 0.7757 | 0.1627 | 0.1580 | 0.1965 |
| w/ Transformer | 0.7962 | 0.6618 | 0.1326 | 0.1489 | 0.1884 |
| w/o LGFF | <u>0.8178</u> | <u>0.6470</u> | <u>0.1212</u> | <u>0.1387</u> | <u>0.1770</u> |
| UDFE | **0.8200** | **0.6084** | **0.1070** | **0.1353** | **0.1732** |

TABLE IV
QUANTITATIVE COMPARISON RESULTS OF ABLATION STUDY ON THE BENEFIT OF UDIA BLOCK. THE VALUE IN BOLD AND UNDERLINE MEANS THE BEST AND SUB-OPTIMAL RESULTS RESPECTIVELY

| Variant | $\delta_3$ | Abs Rel | Sq Rel | RMSE | log10 |
|---|---|---|---|---|---|
| w/o UDIA | 0.8062 | 0.7073 | 0.1368 | 0.1460 | 0.1850 |
| w HEF | 0.8088 | <u>0.5982</u> | <u>0.1123</u> | 0.1415 | 0.1805 |
| w MEM | <u>0.8153</u> | 0.6107 | 0.1185 | <u>0.1370</u> | <u>0.1777</u> |
| UDIA | **0.8200** | **0.6084** | **0.1070** | **0.1353** | **0.1732** |

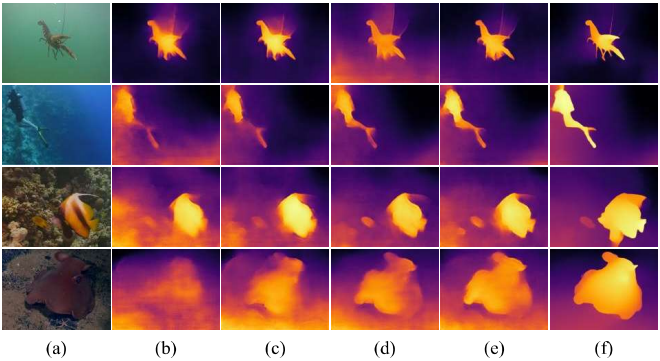

(a)  (b)  (c)  (d)  (e)  (f)

Fig. 9. Qualitative comparison results of ablation study on the benefit of UDFE block. (a) RGB. (b) w/ CNN. (c) w/ Transformer. (d) w/o LGFF. (e) UDFE. (f) Ground Truth.

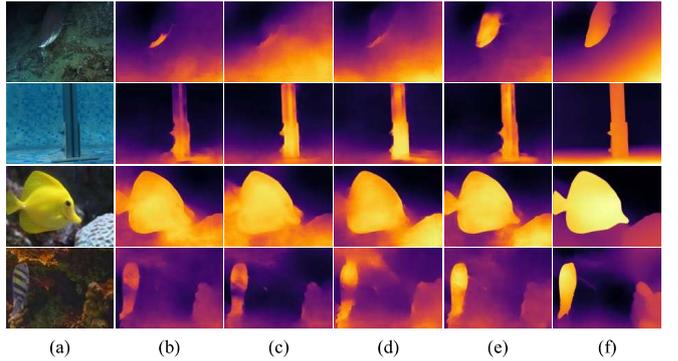

(a)  (b)  (c)  (d)  (e)  (f)

Fig. 10. Qualitative comparison results of ablation study on the benefit of UDFE block. (a) RGB. (b) w/o UDIA. (c) w/ HEF. (d) w/ MTM. (e) UDIA. (f) Ground Truth.

*2) The benefit of UDIA module:* The UDIA module is designed for aggregating the medium transmission maps and hierarchical encoded features to guide the decoding process. To evaluate the importance of UDIA, we establish several variants to substitute UDIA for providing guidance information, which are as follows

- w/o UDIA denotes no information serve as guidance for the decoding process.
- w/ HEF denotes the hierarchical encoded features serve as guidance for the decoding process.
- w/ MTM denotes the medium transmission maps serve as guidance for the decoding process.

The experiments results are presented in Fig. 10 and Table IV. As shown in Fig. 10, the ablated designs 'w/o UDIA' produces poor results with discontinuous depth due to the lack of guidance information. 'w/ HEF' and 'w/ MTM' introduce single guiding information from the deep neural network and physical model, but it is not sufficient to produce satisfactory results. It should be noted that the aforementioned designs may introduce the gradient explosion problem during the training process, leading to difficulties for the model to converge. Compared to all the ablated designs, The benefit of introducing the UDIA module is to enhance depth information by computing the complementation between the hierarchical encoded features and medium transmission maps. The enhanced features, serving as the guidance with depth information from the physical model and deep neural network for the decoding process, is beneficial for achieving a comprehensive understanding of underwater scenes, thus leading to the optimal depth estimation results.

### F. Complexity Analysis

Complexity comparisons experiments are conducted to verify the efficiency of the proposed UMono, while models' parameters, memory and FLOPs served as the evaluation metrics. As shown in Table V, LiteMono [58] and UDepth [37] achieve superior model size(8.78M) and FLOPs(77.16G) respectively, however the balance between complexity and precision is not satisfactory. LapDepth [56] and TransDepth [59] yield sub-optimal quantitative comparisons results, while have a large model size with over $2.6\times$ and $8.8\times$ parameters than UMono. Compared with these methods, UMono's performance in model complexity is less than excellent. However, in the aforementioned qualitative and quantitative analysis, UMono achieves a promising improvement in all evaluation metrics. In general, we hold the opinion that these gaps on complexity are acceptable under the premise of relatively high precision.

TABLE V
COMPARISON RESULTS FOR COMPLEXITY OF THE PROPOSED UMONO AND COMPARED METHODS. THE INPUT SIZE IS SET TO $640 \times 480$ FOR CALCULATING

| Model | Params (M) | Memory (GB) | FLOPs (G) |
|---|---|---|---|
| BTS | 16.34 | 62.65 | 246.80 |
| LapDepth | 73.57 | 281.62 | 364.57 |
| LiteMono | 8.78 | 33.60 | 112.12 |
| TransDepth | 247.40 | 944.13 | 1187.67 |
| UDepth | 15.65 | 59.78 | 77.16 |
| UMono | 28.11 | 107.61 | 148.91 |

## V. CONCLUSION

Generally, we propose a deep learning-based supervised framework for underwater monocular depth estimation. In the hybrid encoder, we combine the CNN and Transformer to fully extract the local and global information, where the following LGFF module effectively integrate these features to enhance the feature representation. In addition, we incorporate the medium transmission map as the underwater domain knowledge into the network by a designed cross attention mechanism, and utilize the UDIA module to aggregate the depth-related information as the guidance for decoding process. Extensive experimental results demonstrate that the proposed UMono outperforms existing methods in both visual quality and quantitative metrics and the designed module is effective for underwater monocular depth estimation. In the future, we plan to joint underwater monocular depth estimation with other visual tasks, such as underwater image enhancement and underwater object detection, leveraging the relevance of these visual tasks to achieve better performance.

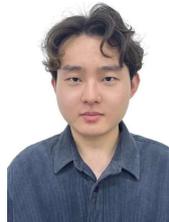

**Jian Wang** received the B.S. degree in communication engineering from North China Electric Power University, Baoding, China, in 2022. He is currently pursuing the M.S. degree with the School of Information Science and Engineering, Ocean University of China, Qingdao, China.

His current research interests include underwater image processing, deep learning, and machine learning.

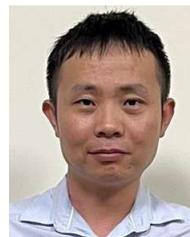

**Jing Wang** received the bachelor's degree in communication engineering from the Tianjin University of Science and Technology, Tianjin, China, in 2010, the master's degree in signal and information processing from the University of Chinese Academy of Sciences, Beijing, China, in 2013, and the Ph.D. degree from Griffith University, Nathan, QLD, Australia in 2020. He was a postdoctoral research fellow with the Commonwealth Scientific and Industrial Research Organisation, Canberra, Australia from 2020 to 2022.

He is currently a research scientist in the Queensland Government, an adjunct research fellow in Griffith University, and a visiting scientist in CSIRO. His research interests include machine learning, computer vision, image processing, medical imaging, and remote sensing.

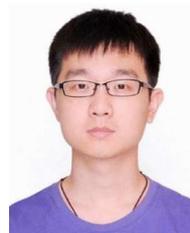

**Shenghui Rong** received the B.S. degree in electronic science and technology and the Ph.D. degree in physical electronics from Xidian University, Xi'an, China, in 2011 and 2018, respectively.

In 2016, he was funded by the China Scholarship Council (CSC) to conduct his research in the direction of 3-D image processing and recognition at Griffith University, Brisbane, Australia. He is currently an Associate professor with the School of Information Science and Engineering,Ocean University of China, Qingdao, China. His current research interests include optoelectronic countermeasures, computer vision, and pattern recognition.

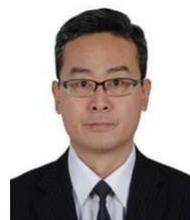

**Bo He** (Member, IEEE) received the M.S. degree in inertial technology and navigation equipment and Ph.D. degree in control theory and control engineering from the Harbin Institute of Technology, Harbin, China, in 1996 and 1999, respectively.

He is currently a Full Professor with the Ocean University of China, Qingdao, China. From 2000 to 2003, he was a postdoctoral fellow with Nanyang Technological University, Singapore. His research work focused on the precise navigation, control, and communication for the platform of mobile robots and unmanned vehicles. His current research interests include image analysis, autonomous underwater vehicle design and applications, and machine learning